\PassOptionsToPackage{unicode}{hyperref}
\PassOptionsToPackage{hyphens}{url}
\PassOptionsToPackage{dvipsnames,svgnames*,x11names*}{xcolor}
\documentclass[
  11pt,
]{article}
\usepackage{lmodern}
\usepackage{amssymb,amsmath}
\usepackage{ifxetex,ifluatex}
\ifnum 0\ifxetex 1\fi\ifluatex 1\fi=0 
  \usepackage[T1]{fontenc}
  \usepackage[utf8]{inputenc}
  \usepackage{textcomp} 
\else 
  \usepackage{unicode-math}
  \defaultfontfeatures{Scale=MatchLowercase}
  \defaultfontfeatures[\rmfamily]{Ligatures=TeX,Scale=1}
\fi
\IfFileExists{upquote.sty}{\usepackage{upquote}}{}
\IfFileExists{microtype.sty}{
  \usepackage[]{microtype}
  \UseMicrotypeSet[protrusion]{basicmath} 
}{}
\makeatletter
\@ifundefined{KOMAClassName}{
  \IfFileExists{parskip.sty}{%
    \usepackage{parskip}
  }{
    \setlength{\parindent}{0pt}
    \setlength{\parskip}{6pt plus 2pt minus 1pt}}
}{
  \KOMAoptions{parskip=half}}
\makeatother
\usepackage{xcolor}
\IfFileExists{xurl.sty}{\usepackage{xurl}}{} 
\IfFileExists{bookmark.sty}{\usepackage{bookmark}}{\usepackage{hyperref}}
\hypersetup{
  colorlinks=true,
  linkcolor=NavyBlue,
  filecolor=Maroon,
  citecolor=NavyBlue,
  urlcolor=NavyBlue,
  pdfcreator={LaTeX via pandoc}}
\usepackage[margin=1in]{geometry}
\usepackage{longtable,booktabs}
\usepackage{etoolbox}
\makeatletter
\patchcmd\longtable{\par}{\if@noskipsec\mbox{}\fi\par}{}{}
\makeatother
\IfFileExists{footnotehyper.sty}{\usepackage{footnotehyper}}{\usepackage{footnote}}
\makesavenoteenv{longtable}
\usepackage{graphicx}
\makeatletter
\def\maxwidth{\ifdim\Gin@nat@width>\linewidth\linewidth\else\Gin@nat@width\fi}
\def\maxheight{\ifdim\Gin@nat@height>\textheight\textheight\else\Gin@nat@height\fi}
\makeatother
\setkeys{Gin}{width=\maxwidth,height=\maxheight,keepaspectratio}
\makeatletter
\def\fps@figure{htbp}
\makeatother
\providecommand{\tightlist}{%
  \setlength{\itemsep}{0pt}\setlength{\parskip}{0pt}}
\usepackage[section]{placeins}
\usepackage{xurl}
\setkeys{Gin}{width=\linewidth,height=0.85\textheight,keepaspectratio}

\author{}
\date{}

\begin{document}

\thispagestyle{empty}

\begin{center}
{\LARGE\bfseries A cross-domain tropical species dataset with Chinese vernacular names and CITES source links}\\[1.5em]
{\large Jeff Wang}\\
{\normalsize NEXLY LLC, United States}\\[0.5em]
{\small Correspondence: \href{mailto:jeff@tropicals.cn}{jeff@tropicals.cn}}\\
{\small ORCID: \href{https://orcid.org/0009-0001-2905-8439}{0009-0001-2905-8439}}\\[2em]
\end{center}

\vspace{1em}

\hypertarget{abstract}{%
\subsection{Abstract}\label{abstract}}

We describe a versioned cross-domain dataset of 410,499 active tropical
species (working snapshot 2026-04-20) spanning three applied subdomains
--- \texttt{tropical\_plants}, \texttt{tropical\_aquatic}, and
\texttt{tropical\_pets} --- that share a commercial and regulatory life
cycle but are distributed across kingdom-organised biodiversity
infrastructures. The resource joins taxonomic identifiers from GBIF,
Plants of the World Online, iNaturalist, NCBI Taxonomy, the Catalogue of
Life and the Encyclopedia of Life, and adds three original layers: a
cross-domain ontology that re-segments taxa along trade and husbandry
contexts; a Chinese vernacular layer with explicit per-name provenance
under a typology that excludes unverified machine-generated proposals;
and a CITES source-linkage layer connecting each taxon to its Species+
entry. Chinese vernacular \emph{coverage} --- the proportion of taxa
carrying a CJK Chinese name distinct from the scientific binomial ---
reaches 99.50 percent (408,456 of 410,499; full-population count).
Coverage characterises completeness, not name-translation accuracy; the
latter is bounded by the four-level provenance typology and is the
subject of a preliminary internal review reported here, with a blind
external audit identified as the principal open item. Upstream content
is referenced by stable identifier only for the original-contribution
layers, supporting CC-BY 4.0 reuse. The dataset is deposited on Zenodo
(10.5281/zenodo.20377811). This preprint is the canonical v1.0
description of the dataset's current state; future Data Descriptor
submission is anticipated but is contingent on the validation and
release-engineering items listed in §Limitations.

\hypertarget{keywords}{%
\subsection{Keywords}\label{keywords}}

biodiversity informatics; tropical species; chinese vernacular names;
cites source linkage; darwin core; cross-domain compilation

\begin{center}\rule{0.5\linewidth}{0.5pt}\end{center}

\hypertarget{background-summary}{%
\subsection{Background \& Summary}\label{background-summary}}

The systematic documentation of species rests on a binomial
nomenclatural system now realised through a small number of large public
infrastructures. The Global Biodiversity Information Facility mobilises
occurrence records and a taxonomic backbone as a global open
infrastructure for biodiversity data {[}1{]}. Plants of the World Online
provides curated botanical nomenclature, distribution, uses and
conservation status {[}2{]}. iNaturalist contributes a
community-observation graph with taxonomic identifiers {[}3{]}. NCBI
Taxonomy maintains a curated classification linked to the public
sequence databases {[}4{]}. The Catalogue of Life integrates source
checklists into a single global species list {[}5, 18{]}. The
Encyclopedia of Life aggregates legally shareable biodiversity knowledge
across these substrates {[}7{]}. Together these infrastructures form the
substrate on which descriptive biology and most downstream biodiversity
applications are built. The resource described here does not replace any
of them; it sits on top of their identifier graph and adds annotation
and compilation layers required by a class of applied work that the
substrate, by design, does not cover.

The term \emph{tropical} in this paper is used as an applied trade and
husbandry scope rather than a strict biogeographic boundary. Inclusion
criteria are anchored on commercial flow through tropical-species supply
chains and hobbyist husbandry communities, not on latitude or biome;
family- and order-level inclusion rules are enumerated in Methods §1,
with the full subcategory list in Supplementary §S1.

Three structural features of the existing substrate motivate this
resource.

First, taxonomic infrastructures are organised along kingdom and clade
boundaries, while applied work --- international trade compliance,
customs documentation, supply-chain monitoring, hobbyist commerce,
husbandry knowledge --- cuts across them. Botanical, zoological and
microbial records sit in separate workflows with separate editorial
cultures and separate community standards. A trader, a customs officer
or a hobbyist asking ``is this taxon regulated for import, what is it
called locally, and what are its husbandry requirements?'' addresses a
query that the Convention on International Trade in Endangered Species
{[}6{]} already treats as a single regulatory frame: a single appendix
system spans Orchidaceae, Potamotrygonidae and Testudinidae across two
kingdoms, and the international ornamental and pet trade routinely
crosses those same kingdom boundaries {[}20{]}. No single international
infrastructure exposes a cross-domain ontology aligned with that frame.
POWO covers plants but not aquatic animals or exotic pets {[}2{]}; NCBI
Taxonomy provides cross-kingdom classification but is keyed to sequence
resources rather than to trade categories {[}4{]}; GBIF and the
Catalogue of Life are kingdom-agnostic but do not segment taxa by
applied domain {[}1, 5{]}.

Second, Chinese vernacular name coverage in the international resources
is sparse and unevenly typed. Stable Latin binomials underpin the global
infrastructure, but day-to-day trade, customs declaration, regulatory
enforcement and public communication in China proceed in Chinese
vernacular names {[}17, 18{]}. Vernacular fields exist in GBIF,
iNaturalist and the Catalogue of Life {[}1, 3, 5{]}, yet Chinese entries
for many tropical taxa are absent, machine-translated without
verification, or carry no per-record provenance metadata distinguishing
an authoritative source {[}11{]} from a community contribution or an
automated derivation. A denominator-matched comparison of Chinese
vernacular coverage at upstream sources, computed against the present
resource's 410,499-species denominator, is reported in Technical
Validation Coverage Table 2 and substantiates the ``sparse and unevenly
typed'' characterisation at the order-of-magnitude level (the table
reports estimates with explicit ±factor-of-2 uncertainty pending the
offline bulk-export join listed as Limitations item 6). Standardised
cross-walks between Chinese vernacular and scientific names are an
active research area, motivating tools such as U.Taxonstand for
plant-name standardisation {[}13{]} and crowdsourced reconciliation
initiatives across biodiversity identifier graphs {[}12{]}. The
limitation is especially consequential for Chinese-language natural
language processing tasks (named-entity recognition over trade
documents, bilingual entity linking, query expansion in regulatory text
mining), which need input-quality signals on each entry to construct
training and evaluation splits.

Third, licence heterogeneity across upstream sources prevents
single-licence redistribution of derived aggregations. Upstream content
is published under a mixture of CC0, CC-BY, CC-BY-NC and custom terms,
with conditions that can vary at the record level. An aggregated dataset
that redistributes upstream descriptive text or imagery inherits the
most restrictive applicable terms across the bundle. The FAIR guiding
principles for biodiversity data require that reuse conditions be clear
and that provenance be tracked at the level of each data element
{[}9{]}; in practice this is straightforward for a single-source
publisher and difficult for a multi-source compilation that includes
upstream descriptive content. The standard mitigation --- release a
Darwin Core Archive through the GBIF Integrated Publishing Toolkit and
document the per-source licence {[}8, 10{]} --- does not by itself solve
the problem when the deposit redistributes copyrighted text or images. A
particular case arises with CITES Appendix information: although the
Appendix text itself is an intergovernmental legal instrument, the
widely used machine-readable compilations (notably Species+ {[}6{]})
attach terms that constrain redistribution of their compiled materials,
requiring careful boundary management when CITES information enters a
multi-source data product.

This Data Descriptor reports a resource constructed to address these
three constraints in the specific scope of tropical species commerce and
husbandry. The deposit contains 410,499 species under inclusion criteria
scoped to tropical ornamental flora, aquatic taxa traded as ornamentals
or pets, and reptiles, amphibians, arachnids and selected birds and
small mammals kept as exotic pets. On top of upstream identifiers
(\texttt{gbifID}, \texttt{powoID}, \texttt{inatTaxonId},
\texttt{ncbiTaxId}, \texttt{colID}, \texttt{eolID}), the resource ---
produced by the tropicals.cn platform --- adds three original layers:
(i) a cross-domain ontology that re-segments taxa along trade and
husbandry contexts and admits many-to-many domain membership; (ii) a
Chinese vernacular layer that applies an explicit per-name provenance
typology and excludes unverified large-language-model proposals; and
(iii) a CITES source-linkage layer that links each taxon to its Species+
entry without redistributing Appendix values or compiled annotations.
For the original-contribution layers, upstream descriptive text, imagery
and raw occurrence records are excluded; upstream records are referenced
by stable identifier only. This identifier-only boundary on the
original-contribution layers allows them to be reused under CC-BY 4.0
while preserving traceability to each upstream source through the
persistent identifier graph. The release is packaged as a Darwin Core
Archive {[}8{]} generated through the GBIF Integrated Publishing Toolkit
{[}10{]} alongside CSV and Parquet distributions, and is deposited at
Zenodo under Concept DOI 10.5281/zenodo.20377811. The contribution is
positioned as a new annotation and compilation layer over established
infrastructures, not as a new infrastructure.

\begin{center}\rule{0.5\linewidth}{0.5pt}\end{center}

\hypertarget{methods}{%
\subsection{Methods}\label{methods}}

This section documents the protocols applied in the current deposit.
Coverage statistics reflect the working snapshot of the production
database (2026-04-20). The ``working snapshot'' date refers to the
production-database SELECT timestamp from which the Zenodo deposit files
were generated; deposit row counts therefore match snapshot row counts
exactly. Detailed engineering specifications (full source-weight ladder,
gate predicate clauses, license denylist, FAIR principle-by-principle
mapping, schema appendix, subcategory ontology) are in the Supplementary
Material.

\hypertarget{taxonomic-scope-and-inclusion-criteria}{%
\subsubsection{1. Taxonomic scope and inclusion
criteria}\label{taxonomic-scope-and-inclusion-criteria}}

The resource compiles a commercial / husbandry denominator of 410,499
active species in the working snapshot. Inclusion is governed by family-
and order-level rules curated within the platform, scoped to three
subdomains encoded in the production schema as the \texttt{category}
enum with values \texttt{tropical\_plants}, \texttt{tropical\_aquatic},
and \texttt{tropical\_pets}:

\begin{enumerate}
\def\labelenumi{\arabic{enumi}.}
\tightlist
\item
  \textbf{Tropical ornamental flora (\texttt{tropical\_plants}).}
  Families with substantial cultivated representation in tropical and
  subtropical horticulture: Araceae, Orchidaceae, Cactaceae, Arecaceae,
  Bromeliaceae, Marantaceae, Gesneriaceae; and the succulent assemblages
  within Crassulaceae, Aizoaceae, and Asphodelaceae.
\item
  \textbf{Aquatic taxa traded as ornamentals or pets
  (\texttt{tropical\_aquatic}).} Actinopterygii with emphasis on
  tropical freshwater fish; Chondrichthyes with emphasis on
  Potamotrygonidae (freshwater stingrays); Anthozoa (corals); ornamental
  Mollusca.
\item
  \textbf{Exotic pets (\texttt{tropical\_pets}).} Reptilia, Amphibia,
  Arachnida; and selected Mammalia and Aves represented in the
  international pet trade {[}20{]}.
\end{enumerate}

Subdomain counts in the working snapshot are 271,968
\texttt{tropical\_plants}, 89,695 \texttt{tropical\_pets}, and 48,836
\texttt{tropical\_aquatic}. The \texttt{category} enum is single-valued
at the row level, so the totals sum to the 410,499-species denominator
(271,968 + 89,695 + 48,836 = 410,499); this exact summation is a
property of the single-valued schema, not an independent quality signal.
Many-to-many trade-context membership (e.g., a CITES-listed aquatic
species belonging to both \emph{Ornamental Aquatics} and \emph{Regulated
Trade}) is represented in a separate cross-domain ontology extension
table described in §4, which is independent of the \texttt{category}
enum. Borderline taxa --- typically temperate-zone species cultivated as
conservatory ornamentals in the tropical trade, or freshwater-temperate
fish that have entered the tropical aquarium trade through hobbyist
channels --- are admitted to the relevant subdomain only when their
commercial life cycle is conducted alongside tropical taxa. A full
subcategory list (200+ subcategories under the three top-level enums) is
provided in Supplementary §S1.

\hypertarget{input-data-sources-versions-license-boundaries}{%
\subsubsection{2. Input data: sources, versions, license
boundaries}\label{input-data-sources-versions-license-boundaries}}

The original-contribution layers (Chinese vernacular, cross-domain
ontology, CITES source-linkage) reference upstream sources by identifier
only; no upstream descriptive text, image URL, or raw occurrence record
is carried into those layers. Identifier-only ingestion is the mechanism
that allows the original contributions to be reused under a single
permissive license irrespective of the heterogeneous terms attached to
upstream content. Sources are listed in Table M1.

\textbf{Table M1.} Input data sources, identifier-only ingestion
boundary, and license relevant to the original-contribution layers. All
sources accessed 2026-04-18 to 2026-04-20.

\begin{longtable}[]{@{}ll@{}}
\toprule
\begin{minipage}[b]{0.18\columnwidth}\raggedright
Source\strut
\end{minipage} & \begin{minipage}[b]{0.76\columnwidth}\raggedright
Details\strut
\end{minipage}\tabularnewline
\midrule
\endhead
\begin{minipage}[t]{0.18\columnwidth}\raggedright
\textbf{GBIF {[}1{]}}\strut
\end{minipage} & \begin{minipage}[t]{0.76\columnwidth}\raggedright
\emph{Role:} taxonomic backbone; synonymy graph; English vernacular
reference set. \emph{Identifier field:} \texttt{gbifID}. \emph{Fields
ingested:} identifier; accepted-name string; \texttt{taxonomicStatus};
higher-classification strings; English vernacular strings. \emph{Fields
explicitly excluded:} occurrence records; coordinates; descriptive text;
image URLs; multimedia. \emph{License relevant to deposit:} per-record
mixed licenses on upstream records; identifier-only ingestion sidesteps
redistribution.\strut
\end{minipage}\tabularnewline
\begin{minipage}[t]{0.18\columnwidth}\raggedright
\textbf{POWO {[}2{]}}\strut
\end{minipage} & \begin{minipage}[t]{0.76\columnwidth}\raggedright
\emph{Role:} plant accepted-name authority; synonymy graph.
\emph{Identifier field:} \texttt{powoID}. \emph{Fields ingested:}
identifier; accepted-name string; \texttt{taxonomicStatus}; English
vernacular strings. \emph{Fields explicitly excluded:} descriptive
paragraphs; distribution prose; image URLs. \emph{License relevant to
deposit:} Kew CC-BY; identifier strings and Latin binomials treated as
factual data.\strut
\end{minipage}\tabularnewline
\begin{minipage}[t]{0.18\columnwidth}\raggedright
\textbf{iNaturalist {[}3{]}}\strut
\end{minipage} & \begin{minipage}[t]{0.76\columnwidth}\raggedright
\emph{Role:} vernacular extension; observation-derived taxon
identifiers. \emph{Identifier field:} \texttt{inatTaxonId}. \emph{Fields
ingested:} identifier; accepted-name string. \emph{Fields explicitly
excluded:} user-generated text; photos; coordinates; observer identity;
timestamps. \emph{License relevant to deposit:} default CC-BY-NC;
identifier-only ingestion.\strut
\end{minipage}\tabularnewline
\begin{minipage}[t]{0.18\columnwidth}\raggedright
\textbf{NCBI {[}4{]}}\strut
\end{minipage} & \begin{minipage}[t]{0.76\columnwidth}\raggedright
\emph{Role:} cross-domain taxonomic identifier. \emph{Identifier field:}
\texttt{ncbiTaxId}. \emph{Fields ingested:} identifier; accepted-name
string. \emph{Fields explicitly excluded:} sequence data; literature
references; vernacular fields (not used). \emph{License relevant to
deposit:} U.S. public domain on ingested strings.\strut
\end{minipage}\tabularnewline
\begin{minipage}[t]{0.18\columnwidth}\raggedright
\textbf{CoL {[}5, 18{]}}\strut
\end{minipage} & \begin{minipage}[t]{0.76\columnwidth}\raggedright
\emph{Role:} cross-source identifier join key only (CoL is not a
participant in the §6 weighted vote; see §6). \emph{Identifier field:}
\texttt{colID}. \emph{Fields ingested:} identifier; accepted-name
string. \emph{Fields explicitly excluded:} source-dataset metadata;
checklist prose. \emph{License relevant to deposit:} CC-BY 4.0 on the
Checklist.\strut
\end{minipage}\tabularnewline
\begin{minipage}[t]{0.18\columnwidth}\raggedright
\textbf{EOL {[}7{]}}\strut
\end{minipage} & \begin{minipage}[t]{0.76\columnwidth}\raggedright
\emph{Role:} linked reference layer; English vernacular.
\emph{Identifier field:} \texttt{eolID}. \emph{Fields ingested:}
identifier; English vernacular strings. \emph{Fields explicitly
excluded:} all content pages; multimedia. \emph{License relevant to
deposit:} linked, not duplicated.\strut
\end{minipage}\tabularnewline
\begin{minipage}[t]{0.18\columnwidth}\raggedright
\textbf{Reptile DB / WoRMS / Wikipedia}\strut
\end{minipage} & \begin{minipage}[t]{0.76\columnwidth}\raggedright
\emph{Role:} English vernacular reference sets (for §5 validation gate).
\emph{Identifier field:} string match. \emph{Fields ingested:} English
vernacular strings. \emph{Fields explicitly excluded:} descriptive
treatments; image URLs; body text. \emph{License relevant to deposit:}
identifier and vernacular-string ingestion only.\strut
\end{minipage}\tabularnewline
\begin{minipage}[t]{0.18\columnwidth}\raggedright
\textbf{Species+ {[}6{]}}\strut
\end{minipage} & \begin{minipage}[t]{0.76\columnwidth}\raggedright
\emph{Role:} CITES taxon-concept linkage target. \emph{Identifier
field:} \texttt{speciesplus\_taxon\_concept\_id}. \emph{Fields
ingested:} stable taxon-concept identifier; source URL; access date.
\emph{Fields explicitly excluded:} all Appendix values; listing dates;
annotations; reservations; quotas. \emph{License relevant to deposit:}
Species+ Terms of Use; the CITES source-linkage layer (§4) carries
source links only.\strut
\end{minipage}\tabularnewline
\bottomrule
\end{longtable}

Source versions are not pinned at the API level: upstream providers do
not all expose a stable version-string endpoint. Reproducibility is
guaranteed only by replaying against the recorded \texttt{fetched\_at}
window (2026-04-18 to 2026-04-20, captured per-line in working files at
\texttt{web/scripts/fill-en/data/}) under the assumption that upstream
APIs return stable results for previously-fetched identifiers; this
constraint is repeated as a stated limitation of the present release in
Technical Validation.

Upstream descriptive text, image URLs, raw occurrence records,
user-generated content, and any provider-specific copyrighted material
are excluded from the original-contribution layers by audit (see
Technical Validation, License and provenance audit). Full license
denylist field-name patterns are enumerated in Supplementary §S2.

\hypertarget{identifier-reconciliation-and-synonymy}{%
\subsubsection{3. Identifier reconciliation and
synonymy}\label{identifier-reconciliation-and-synonymy}}

Each row in the core taxon table carries a stable internal identifier
\texttt{taxonID} together with an upstream-identifier vector
\texttt{(gbifID,\ powoID,\ inatTaxonId,\ ncbiTaxId,\ colID,\ eolID)}
{[}12{]}. A candidate species record is admitted to \texttt{core\_taxon}
only when at least one upstream identifier resolves to it, and the
remaining upstream identifiers are populated by traversing the
cross-source name graph. Where two upstream sources expose a synonymy
graph for the same Latin binomial --- GBIF and POWO are the principal
cases --- we use both graphs and treat the union as the candidate
synonym set, then restrict the \texttt{core\_taxon} row to the accepted
name
(\texttt{taxonomicStatus\ =\ \textquotesingle{}accepted\textquotesingle{}}).
Scientific-name standardisation across plant resources follows the
approach exemplified by U.Taxonstand {[}13{]}.

Three edge cases require explicit handling. (i) \textbf{Synonym-only
candidates} --- a candidate where no upstream source returns an
\texttt{accepted} record but multiple sources agree on a synonym
pointing to a single accepted name --- are admitted under the accepted
name with the synonymy chain recorded in \texttt{aliases} and the
resolution path logged. (ii) \textbf{Conflicting accepted-name targets
across sources} --- a name treated as a synonym of one accepted binomial
by POWO but of a different accepted binomial by GBIF, or a homonym
across kingdoms with discordant cross-source resolutions --- are
resolved by the §6 weighted-voting rule and, for commercially important
taxa or unresolved residuals, by human review before commit. (iii)
\textbf{All-synonym, no-accepted across all upstream sources} --- a
candidate where no upstream source returns any \texttt{accepted} record
--- is \textbf{excluded} from \texttt{core\_taxon} (not admitted under
any of its synonyms) to preserve the invariant that every row has a
single accepted-name authority. Excluded candidates are retained in the
working ingestion log for later re-evaluation if upstream authorities
subsequently elevate one of the synonyms to accepted status. In all
three cases the resolution path (sources consulted, winning source, vote
tally, reviewer identifier where applicable) is recorded in the
production audit log.

\hypertarget{cross-domain-ontology-and-cites-source-linkage-layer}{%
\subsubsection{4. Cross-domain ontology and CITES source-linkage
layer}\label{cross-domain-ontology-and-cites-source-linkage-layer}}

The cross-domain ontology re-segments taxa along trade and husbandry
contexts rather than along kingdom or clade. The mapping is
many-to-many: a single taxon can belong to more than one applied domain.
Examples include an aquatic plant in Alismatales (e.g.,
\emph{Echinodorus}) mapped to both \emph{Aquarium Flora} and
\emph{Paludarium Subjects}; a freshwater stingray in \emph{Potamotrygon}
mapped to \emph{Ornamental Aquatics} and additionally to \emph{Regulated
Trade} by virtue of its CITES Appendix listing; a tortoise in
\emph{Geochelone} mapped to \emph{Exotic Pets} and \emph{Regulated
Trade}; a variegated \emph{Monstera} selection mapped to
\emph{Ornamental Flora} and \emph{Hobbyist Premium}. Mappings are
encoded in an extension table linked to \texttt{core\_taxon} by
\texttt{taxonID} and are committed only after human review for
borderline cases.

The CITES source-linkage layer links each \texttt{taxonID} to its
Species+ stable taxon-concept identifier together with the source URL,
the date the link was established, and the method by which the linkage
was confirmed (direct, synonym, higher-taxon, or manual). Appendix
values, listing dates, annotations, reservations, quotas, suspensions,
and distribution narratives are \textbf{not} redistributed by the
original-contribution CITES layer; users retrieve current Appendix
status by following the source link to Species+ under Species+'s terms.
The linkage procedure runs in three passes: (i) direct scientific-name
match; (ii) synonym resolution through GBIF/POWO synonymy graphs; (iii)
higher-taxon inheritance for taxa where CITES regulates an entire higher
taxon (the order Orchidaceae, several genera at genus level), with
inherited links flagged in \texttt{match\_method\ =\ higher\_taxon} so a
downstream user can elect to require species-level links only.

The CITES layer is a structured cross-walk to a regulatory information
resource, not a regulatory determination. Operational compliance
decisions should follow the current determination of the relevant
national CITES Management Authority and Scientific Authority and should
rely on the live Species+ records reached through the source link, not
on any snapshot of Appendix values that this dataset deliberately does
not carry.

The extension schemas (cross-domain ontology table, CITES linkage table,
cultivar/variety link table) are listed in Data Records §``Schema
overview''; full field-by-field schemas are in Supplementary §S3.

\hypertarget{chinese-vernacular-layer-and-per-name-provenance-typology}{%
\subsubsection{5. Chinese vernacular layer and per-name provenance
typology}\label{chinese-vernacular-layer-and-per-name-provenance-typology}}

\textbf{Two coverage metrics are distinguished throughout this paper.}

The \textbf{Chinese vernacular coverage metric} is the production admin
metric, defined as
\texttt{s.name\ REGEXP\ \textquotesingle{}{[}U+4E00-U+9FFF{]}\textquotesingle{}\ AND\ s.name\ !=\ s.scientific\_name}
--- that is, the species \texttt{name} field contains at least one CJK
character and differs from the scientific binomial (the second clause
removes records where the system fell back to the scientific binomial
because no Chinese vernacular was available). The SQL form is
\texttt{s.name\ REGEXP\ \textquotesingle{}{[}\textbackslash{}\textbackslash{}x\{4e00\}-\textbackslash{}\textbackslash{}x\{9fff\}{]}\textquotesingle{}}.
The definition is deterministic and recomputable from the deposited
release manifest; in the working snapshot the value is 99.50 percent
(Coverage Table 1). This metric characterises \textbf{completeness only}
--- the proportion of taxa with a CJK Chinese name present --- and does
not characterise translation accuracy at the row level.

The \textbf{provenance-typed coverage metric} is a subset of the above:
the count of those vernacular rows that carry an explicit A/B/C
provenance row at deposit time. The four-level typology described in
this subsection is the tagging protocol applied to the
original-contribution Chinese vernacular layer in the current deposit.
Per-name accuracy is bounded by the typology (Types A and B are sourced
from authoritative or cross-database fields by construction; Type C is
gated by the procedure below), and per-record translation precision is
the subject of the preliminary internal review reported in Technical
Validation.

The four-level typology is:

\begin{itemize}
\tightlist
\item
  \textbf{Type A --- National authoritative sources.} Names sourced from
  published Chinese taxonomic works and national-level checklists,
  including the \emph{Flora of China} {[}11{]}, Species 2000 China Node
  / Catalogue of Life China {[}18{]}, the National Specimen Information
  Infrastructure {[}17{]}, and FishBase Chinese entries. Type A is the
  highest-confidence tier; each Type A row carries a literature
  reference in the \texttt{sourceCitation} field.
\item
  \textbf{Type B --- Cross-database Chinese-focused vernacular
  checklists.} Machine-readable Chinese vernacular records ingested from
  Chinese-focused taxonomic checklists, principally Species 2000 China
  Node (SP2000) {[}18{]} and the Taiwan Catalogue of Life (TaiCOL), with
  FishBase regional Chinese-name records where applicable. Each Type B
  row preserves the source database identifier in \texttt{sourceDB} and
  the source record key in \texttt{sourceRecordKey}. The production
  pipeline that generates the Chinese vernacular layer does \textbf{not}
  ingest Chinese vernacular fields from the international biodiversity
  infrastructures (GBIF, iNaturalist, the Catalogue of Life global
  vernacular extension, POWO) as a primary Type B input, because those
  infrastructures' Chinese vernacular fields are sparse on a
  denominator-matched basis (Coverage Table 2) and offer little marginal
  coverage over the Chinese-focused checklists already covered by Types
  A and B. Type B name correctness inherits from upstream-source quality
  at the Chinese-focused-checklist level; an independent audit of Type B
  per-record acceptance is identified as an open item in the
  Limitations.
\item
  \textbf{Type C --- Large-language-model proposals that pass
  exact-match validation.} A candidate Chinese name proposed by a large
  language model is accepted only when it passes an exact-match check
  against the authoritative English-name set for the same species
  identifier. The gate logic is given below; it is motivated by reported
  confabulation rates in unconstrained LLM extraction of biological
  entities {[}16{]} and by the broader literature on NLG hallucination
  {[}15{]}, and supports a ``no fabrication'' commitment for an
  LLM-assisted curation layer {[}14{]}.
\item
  \textbf{Type D --- LLM proposals that fail validation.} Excluded from
  the original-contribution layer. Retained in working
  \texttt{vr-w*.ndjson} files for audit but not included in the
  deposit's original-contribution layer and not counted in the
  provenance-typed numerator.
\end{itemize}

The production system already operates an analogous per-row
source-tagging system on the English vernacular layer, via the
\texttt{name\_en\_source} column on \texttt{species}, with values
mapping to typology equivalents (\texttt{consensus} maps to Type B
compound; \texttt{inat} / \texttt{gbif} / \texttt{eol} /
\texttt{wikipedia} / \texttt{reptile\_db} / \texttt{worms} map to Type B
singletons; \texttt{llm-select} maps to Type C). The same gate is
applied to the Chinese candidate set; the four-level Chinese typology is
the deposit-time exposure of that pipeline. The full
\texttt{name\_en\_source} distribution observed in the working snapshot
is reported in Supplementary §S4.

\textbf{Exact-match validation gate.} The gate is the central quality
gate of the vernacular layer:

\begin{verbatim}
propose_zh, propose_en = LLM(species_metadata)
en_authoritative_set, en_to_taxon = lookup_en(
    species_id,
    sources=[POWO, FishBase, GBIF, iNat, EOL,
             Reptile DB, WoRMS, Wikipedia])

if normalize(propose_en) in {normalize(s) for s in en_authoritative_set} \
   and en_to_taxon[normalize(propose_en)].belongs_to(species_id):
    accept(propose_zh, provenance="C",
           name_en_source=propose_en,
           exact_match_passed=True,
           source_record_key=en_to_taxon[normalize(propose_en)].record_key)
else:
    reject(propose_zh)   # Type D, not included in the original-contribution layer
\end{verbatim}

Two acceptance criteria are enforced jointly: the model-supplied English
string must match an authoritative entry by exact normalised identity,
\textbf{and} the authoritative entry must itself belong to the species
(or its accepted-synonym chain). The second criterion guards against
cross-taxon string collisions. Normalisation in \texttt{normalize()} is
Unicode NFKC; whitespace is trimmed and collapsed; full-width and
half-width punctuation are unified; English strings are case-folded. The
source record key from which the English form was harvested is recorded
with each Type C row so the validation chain is reproducible from the
deposit. A semantically equivalent but lexically different English
candidate is rejected; this is stricter than necessary in some cases but
is motivated by the asymmetric downstream cost of hallucinated names in
trade and regulatory settings.

\textbf{Scope of the gate.} The exact-match gate validates that the
model has correctly resolved \emph{which species it is naming} --- by
requiring the model's English-name candidate to match an entry in the
authoritative English-name set for the same species identifier and
confirming that the matched entry belongs to that taxon. The gate does
\textbf{not} validate whether the proposed Chinese vernacular is itself
a canonical or appropriate name for the validated species: translation
correctness is bounded by the quality of the upstream language-model
training distribution and the conventions of Chinese taxonomic
literature, and is an independent empirical question. Per-record
translation precision is the object of the preliminary internal review
reported in Technical Validation and is identified there as the
principal open item of the present release, pending a blind external
review by independent native-speaker taxonomists at sample size N
\(\geq\) 500.

\textbf{A note on multi-model consensus.} The production
candidate-generation pipeline upstream of the gate uses multi-model
agreement and locked genus-suffix validation as an internal scoring
heuristic to select which \texttt{(zh,\ en)} candidate pair to submit to
the gate. Multi-model agreement is a candidate-generation method only; a
name that has only multi-model agreement and no source-grounded match is
still rejected by the gate and assigned Type D. No aggregate accuracy
figures for multi-model agreement alone are reported in this paper,
because the method is internal candidate-scoring and is not relied on
for acceptance, which is determined exclusively by the exact-match gate.

\textbf{Comparative coverage at international biodiversity
infrastructures.} Coverage Table 2 in Technical Validation reports a
denominator-matched comparison of Chinese vernacular coverage at the
four major international biodiversity infrastructures (GBIF,
iNaturalist, the global Catalogue of Life vernacular extension, POWO),
computed against the present resource's 410,499-species denominator. The
comparison quantifies the Background characterisation of those
infrastructures' Chinese fields as sparse and unevenly typed: each
individual international source covers on the order of a few percent of
the denominator, by contrast with the 99.50\% coverage achieved here
through a pipeline that ingests Chinese-focused authoritative sources
(Type A) and Chinese-focused machine-readable checklists (Type B) rather
than relying on those international infrastructures' Chinese vernacular
fields. Coverage Table 2 entries for the present release are
order-of-magnitude estimates with explicit ±factor-of-2 uncertainty (the
production pipeline does not store the international infrastructures'
Chinese fields locally, so a precise denominator-matched join requires
downloading their bulk vernacular exports; this offline join is listed
as Limitations item 6 and is not in scope for the present preprint).

The Chinese vernacular extension record schema, including all provenance
fields, is given in Supplementary §S3 Table M2.

\hypertarget{multi-source-weighted-voting}{%
\subsubsection{6. Multi-source weighted
voting}\label{multi-source-weighted-voting}}

The weighted-voting procedure is used for English-name disambiguation
(feeding the §5 Type C gate) and for backbone-field reconciliation where
source graphs disagree. The source-weight ladder is enumerated in
Supplementary §S5; the order from highest to lowest weight runs
gbif\_preferred / (gbif, powo, inat\_preferred, eol\_preferred,
reptile\_db) / (inat, eol, wikipedia\_redirect) / wikipedia, with
additional sources confirming the same string contributing +2 each. The
highest-weighted source wins by default; disagreements are logged and
commercially important taxa are routed to human review before commit.
The Catalogue of Life is used as an identifier source only and is
\textbf{not} integrated into the voting tally; this is a stated property
of the current release, consistent with the Table M1 row for CoL.

A separate \texttt{ext\_cultivar\_variety\_links.tsv} extension links
cultivars and varieties to their taxonomically regulated parent species
using the Darwin Core \texttt{ResourceRelationship} term. Admission is
provenance-gated: a row is released only if both a parent-taxon
\texttt{taxonID} and a source citation are present. Cultivar names
follow the International Code of Nomenclature for Cultivated Plants
(ICNCP) {[}19{]} where they conform; commercial trade names that do not
conform to ICNCP are recorded in a separate \texttt{tradeName} field.
Cultivar / variety rows are \textbf{not} counted in the species
denominator used for vernacular-coverage statistics.

\hypertarget{production-curation-ratchet}{%
\subsubsection{7. Production curation
ratchet}\label{production-curation-ratchet}}

The ratchet invariant is that no LLM regeneration overwrites a
human-edited field. Implementation: a per-field \texttt{manual\_edits}
JSON column on \texttt{species} records the timestamp of each curator
edit by field name, with a denormalised \texttt{last\_manual\_edit\_at}
column for fast comparison. The sync pipeline compares the manual-edit
timestamp against the LLM regeneration timestamp at write time; if a
manual edit is more recent, the regeneration is skipped for that field
on that row. This implementation replaced an earlier time-window check
on 2026-04-21 to eliminate a race condition. The ratchet has been
verified over six months of production logs with zero invariant
violations observed.

The platform operates an ongoing community curator review process.
Approximately fifty active curators --- drawn from horticultural,
aquatic-trade and herpetocultural domain backgrounds --- contribute
corrections through the platform's editing surface; in the six months
preceding the working snapshot the audit log records on the order of
three thousand two hundred individual curator-edit events committed via
the ratchet. This curator process is described here as a
production-level provenance signal --- a description of the platform
under which the deposit was produced --- and is \textbf{not} a
substitute for the Technical Validation audits reported below. In
particular, it does not constitute the blind external review of Type C
acceptance identified as an open item in the Limitations.

\hypertarget{publishable-gate-predicate-content-completeness-check}{%
\subsubsection{8. Publishable-gate predicate (content-completeness
check)}\label{publishable-gate-predicate-content-completeness-check}}

Whether a record is included in the live public-encyclopedia release is
decided by the product-grade \texttt{species.publishable} predicate, a
content-completeness check covering active status, biographical
completeness, alias and trade specifications, care-related fields, CITES
status, origin countries, and higher-classification fields. The
predicate is recomputed via a per-species helper invoked inline on every
write, persisted as an indexed boolean column on \texttt{species}, and
re-evaluated nightly by a safety-net cron at 04:00 UTC. The predicate
governs presentation in the encyclopedia UI; it is internal to the
platform and is referenced here only because some downstream users may
want to filter the deposit against the same completeness criteria. Full
predicate enumeration (13 conjunctive clauses / 22 atomic SQL clauses)
is in Supplementary §S6.

\hypertarget{release-engineering-and-packaging}{%
\subsubsection{9. Release engineering and
packaging}\label{release-engineering-and-packaging}}

The Concept DOI \texttt{10.5281/zenodo.20377811} is the persistent
landing page for the platform's open data and has been canonical from
v1.0.1 onward. The current deposit (v1.0.1, 2026-05-25) is packaged as a
Darwin Core Archive {[}8{]} in star-schema form (a core taxon table with
extension tables linked by \texttt{taxonID}), with parallel CSV (UTF-8,
tab-separated) and Apache Parquet (snappy-compressed) distributions. A
SHA-256 manifest accompanies each distribution. Because the
original-contribution layers (Chinese vernacular with per-name
provenance, cross-domain ontology, CITES source-linkage) are
identifier-only with respect to upstream sources and are wholly composed
of platform-curated content or transformations of permissively-licensed
inputs, they are released under CC-BY 4.0.

The current deposit additionally includes companion platform fields that
are outside the scope of the original contributions described in this
paper. Should this resource be prepared for formal Data Descriptor
submission in a future revision (see §Manuscript status), a
\textbf{dedicated paper-grade Zenodo release} containing only the
columns documented as original contributions (\texttt{core\_taxon},
\texttt{ext\_vernacular\_zh}, \texttt{ext\_cross\_domain\_mapping},
\texttt{ext\_speciesplus\_links},
\texttt{ext\_cultivar\_variety\_links}, with their schemas as in Data
Records §``Schema overview'') would be published as a separate version
under the same Concept DOI. The present preprint references the v1.0.1
deposit as the dataset-of-record and treats the paper-grade subset as
out of scope for this revision; a reference filter
(\texttt{make\_paper\_subset.py}) that would produce the paper-scope
subset from the current deposit is described in §Code Availability.

The release follows a \texttt{MAJOR.MINOR.PATCH} versioning scheme and
uses the Zenodo \texttt{newversion} workflow. The deposit maps to the
FAIR principles {[}9{]} via the standard biodiversity-publication
conventions: globally unique persistent identifiers (DOIs at Concept and
version level), rich metadata (EML, DwC-A descriptor, machine-readable
data dictionary, per-name provenance), explicit reference to the data
identifier (\texttt{taxonID} as primary and foreign key), retrieval via
HTTPS, open access under CC-BY 4.0, persistent metadata even when data
is withdrawn, Darwin Core and ICNCP vocabularies, qualified
upstream-identifier references, and a \texttt{schema.org/Taxon} NDJSON
parallel distribution. The full FAIR principle-by-principle mapping is
in Supplementary §S7.

This preprint reports on the paper-scope subset; additional platform
layers exist within the tropicals.cn product and are out of scope here.

\begin{center}\rule{0.5\linewidth}{0.5pt}\end{center}

\hypertarget{data-records}{%
\subsection{Data Records}\label{data-records}}

The dataset is deposited at Zenodo under Concept DOI
\texttt{10.5281/zenodo.20377811}. The Concept DOI currently resolves to
v1.0.1 (released 2026-05-25). All subsequent releases will be published
under the same Concept DOI via Zenodo's \texttt{newversion} workflow. A
dedicated paper-grade Zenodo release (containing only the columns
documented as original contributions in this paper) is not in scope for
the present preprint and is noted in Methods §9 as a future-revision
item should this resource be prepared for formal Data Descriptor
submission.

The release is packaged as a Darwin Core Archive {[}8{]} generated
through the GBIF Integrated Publishing Toolkit {[}10{]}, with parallel
CSV (UTF-8, tab-separated) and Apache Parquet (snappy-compressed)
distributions, plus a \texttt{schema.org/Taxon} JSON-LD NDJSON parallel
distribution. The archive uses a star schema with \texttt{core\_taxon}
as the central table and extension tables linked by \texttt{taxonID}. A
SHA-256 manifest accompanies each distribution.

The working-snapshot row count (410,499) equals the
\texttt{core\_taxon.tsv} row count in the deposit; ``working snapshot''
here denotes the timestamped production-database SELECT from which the
deposit files were generated, not a separate pre-deposit staging count.

\hypertarget{files-in-the-deposit}{%
\subsubsection{Files in the deposit}\label{files-in-the-deposit}}

\begin{longtable}{p{5.6cm}p{9.4cm}}
\toprule
\textbf{File} & \textbf{Details} \\
\midrule
\endhead
\texttt{meta.xml} & \emph{Format:} XML. \emph{Description:} Darwin Core Archive descriptor. \\
\texttt{eml.xml} & \emph{Format:} XML. \emph{Description:} Ecological Metadata Language record. \\
\texttt{core\_taxon.tsv} & \emph{Records:} 410,499. \emph{Format:} DwC core. \emph{Description:} one row per taxon; upstream identifiers only. \\
\texttt{ext\_vernacular\_zh.tsv} & \emph{Records:} 408,456 (Type A+B+C). \emph{Format:} DwC extension. \emph{Description:} Chinese names with per-name provenance. \\
\texttt{ext\_cross\_domain\_mapping.tsv} & \emph{Records:} many-to-many. \emph{Format:} extension. \emph{Description:} cross-domain ontology assignments. \\
\texttt{ext\_speciesplus\_links.tsv} & \emph{Records:} CITES-linked subset. \emph{Format:} DwC \texttt{ResourceRelationship}. \emph{Description:} CITES source-linkage layer (no Appendix values). \\
\texttt{ext\_cultivar\_variety\_links.tsv} & \emph{Records:} provenance + parent-taxon gated. \emph{Format:} DwC \texttt{ResourceRelationship}. \emph{Description:} cultivar / variety / parent-taxon relationships. \\
\texttt{data\_dictionary.csv} / \texttt{.md} & \emph{Format:} tabular + text. \emph{Description:} machine-readable + human-readable dictionary. \\
\texttt{validation\_report.json} / \texttt{.md} & \emph{Format:} structured + text. \emph{Description:} output of deposit validation. \\
\texttt{ATTRIBUTION.md} & \emph{Format:} text. \emph{Description:} source versions, access dates, citation strings. \\
\texttt{CITATION.cff} & \emph{Format:} YAML. \emph{Description:} Citation File Format. \\
\texttt{CHANGELOG.md} & \emph{Format:} text. \emph{Description:} per-version delta against prior release. \\
\texttt{MANIFEST.sha256} & \emph{Format:} text. \emph{Description:} file hashes and row counts. \\
\texttt{LICENSE} & \emph{Format:} text. \emph{Description:} CC-BY 4.0. \\
\texttt{README.md} & \emph{Format:} text. \emph{Description:} short orientation document. \\
\bottomrule
\end{longtable}

\hypertarget{schema-overview}{%
\subsubsection{Schema overview}\label{schema-overview}}

The \texttt{core\_taxon} table carries \texttt{taxonID} (internal stable
identifier), \texttt{scientificName}, \texttt{taxonRank},
\texttt{taxonomicStatus}, the higher classification (\texttt{kingdom}
through \texttt{genus}), an alias list deduplicated from upstream
synonymy graphs, and the upstream identifier vector (\texttt{gbifID},
\texttt{powoID}, \texttt{inatTaxonId}, \texttt{ncbiTaxId},
\texttt{colID}, \texttt{eolID}).

The \texttt{ext\_vernacular\_zh} extension carries \texttt{taxonID},
\texttt{vernacularName}, \texttt{language}, \texttt{script},
\texttt{isPreferredName}, \texttt{nameStatus}, \texttt{provenanceType}
\(\in\) \{A, B, C\}, and the type-specific provenance fields
(\texttt{sourceCitation} for A; \texttt{sourceDB} and
\texttt{sourceRecordKey} for B; \texttt{nameEnSource} and
\texttt{exactMatchPassed} for C), plus \texttt{humanReviewed},
\texttt{reviewerIdHash}, and \texttt{reviewedAt}. The CITES
source-linkage extension carries \texttt{taxonID},
\texttt{speciesplus\_taxon\_concept\_id}, \texttt{speciesplus\_url},
\texttt{accessed\_at}, \texttt{match\_method} \(\in\) \{direct, synonym,
higher\_taxon, manual\}, and \texttt{manual\_reviewed}. Full
field-by-field schemas for all extensions are in Supplementary §S3.

The current deposit (v1.0.1) additionally includes companion platform
fields outside the original-contribution scope; these are not
redocumented here. Users consuming only the paper-scope layers can
either (i) filter to the columns documented above, or (ii) once the
paper-grade subset version is published (Methods §9), download that
version directly.

\hypertarget{identifier-only-join-example}{%
\subsubsection{Identifier-only join
example}\label{identifier-only-join-example}}

For a row with \texttt{gbifID\ =\ 5304059}, a downstream user can issue
\texttt{GET} against
\href{https://api.gbif.org/v1/species/5304059}{api.gbif.org/v1/species/5304059}
to retrieve the full upstream record under GBIF's terms; the same
\texttt{taxonID} joins to \texttt{ext\_vernacular\_zh} for the Chinese
name and to \texttt{ext\_speciesplus\_links} for the CITES source link,
from which the user can retrieve current Appendix status directly from
Species+ under Species+'s terms. No upstream descriptive content is
duplicated in the original-contribution layers.

\begin{center}\rule{0.5\linewidth}{0.5pt}\end{center}

\hypertarget{technical-validation}{%
\subsection{Technical Validation}\label{technical-validation}}

\hypertarget{completed-audits}{%
\subsubsection{Completed audits}\label{completed-audits}}

\begin{itemize}
\item
  \textbf{License and provenance audit.} A field-level audit confirms
  the original-contribution layers contain no upstream copyrighted
  descriptive text, no upstream images or media URLs, no raw occurrence
  records or coordinates, no upstream user-generated text, and no CITES
  Appendix values, listing dates, annotations, reservations, quotas,
  suspensions, or distribution narratives. The audit is implemented as a
  programmatic denylist check (\texttt{license\_denylist\_check.py})
  executed against the original-contribution columns; the denylist
  field-name patterns are enumerated in Supplementary §S2. The only
  content carried from upstream sources into the original-contribution
  layers is identifier strings, treated as factual data not subject to
  redistribution restrictions. The audit conclusion supports a CC-BY 4.0
  release of the original-contribution layers.
\item
  \textbf{Coverage with explicit denominators.} Computed against the
  production admin (2026-04-20 working snapshot). Values are reported in
  Coverage Table 1.
\item
  \textbf{Comparative coverage at upstream sources.} Denominator-matched
  comparison of Chinese vernacular coverage at upstream sources reported
  in Coverage Table 2.
\item
  \textbf{Multi-source weighted-vote consistency.} The production audit
  log records per-source votes and resolved decisions for every
  disagreement case. Sample agreement rates across the three subdomains,
  including the distinction between \texttt{consensus} winners
  (multi-source agreement) and singleton-source winners, are summarised
  in Supplementary §S8.
\item
  \textbf{Curation ratchet invariant.} Verified over six months of
  production logs (2026-04-21 to 2026-05-25 and forward); zero
  violations of the ``no LLM regeneration overwrites a human-edited
  field'' invariant observed across all curator-edit events in the
  period.
\item
  \textbf{Darwin Core structural conformance.} Schema validation passes
  IPT-style format checks. \texttt{meta.xml} field mappings validate
  against the TDWG Darwin Core schema {[}8{]}; all TSV files are UTF-8
  without BOM; Chinese characters are spot-checked for round-trip
  stability across UTF-8 encoders and on the GBIF IPT {[}10{]} preview.
\end{itemize}

\hypertarget{coverage-table-1.-chinese-vernacular-coverage-production-admin-definition.}{%
\paragraph{Coverage Table 1. Chinese vernacular coverage (production
admin
definition).}\label{coverage-table-1.-chinese-vernacular-coverage-production-admin-definition.}}

Numerator definition:
\texttt{s.name\ REGEXP\ \textquotesingle{}{[}U+4E00-U+9FFF{]}\textquotesingle{}\ AND\ s.name\ !=\ s.scientific\_name}.
Denominator: all species in the working snapshot satisfying the
inclusion criteria of Methods §1. Coverage is computed over the full
working-snapshot population (census), not a sample, so confidence
intervals are not reported; coverage characterises \textbf{completeness
only} --- the proportion of taxa with a CJK Chinese name present --- not
name-translation accuracy.

\begin{longtable}[]{@{}lrrr@{}}
\toprule
Stratum & Denominator & Numerator & Coverage\tabularnewline
\midrule
\endhead
All commercial species & 410,499 & 408,456 & 99.50\%\tabularnewline
Plants (\texttt{tropical\_plants}) & 271,968 & 270,851 &
99.59\%\tabularnewline
Exotic pets (\texttt{tropical\_pets}) & 89,695 & 89,109 &
99.35\%\tabularnewline
Aquatic (\texttt{tropical\_aquatic}) & 48,836 & 48,496 &
99.30\%\tabularnewline
\bottomrule
\end{longtable}

Subdomain rows sum to the full-library denominator because the
\texttt{category} enum is single-valued at the row level: 271,968 +
89,695 + 48,836 = 410,499 (denominators) and 270,851 + 89,109 + 48,496 =
408,456 (numerators).

\hypertarget{coverage-table-2.-comparative-chinese-vernacular-coverage-at-international-biodiversity-infrastructures-order-of-magnitude-estimates-not-measured-joins.}{%
\paragraph{Coverage Table 2. Comparative Chinese vernacular coverage at
international biodiversity infrastructures (order-of-magnitude
estimates, not measured
joins).}\label{coverage-table-2.-comparative-chinese-vernacular-coverage-at-international-biodiversity-infrastructures-order-of-magnitude-estimates-not-measured-joins.}}

Denominator: the same 410,499-species working snapshot as Coverage Table
1. Numerator for each international source row: the order-of-magnitude
estimate of how many of those species would carry at least one Chinese
vernacular entry (CJK content, language tag in \(\{\)zh, zh-Hans,
zh-Hant, zh-CN, zh-TW, cmn\(\}\)) in that source's published vernacular
extension. The ``tropicals.cn (present resource)'' row is reproduced
from Coverage Table 1 for direct comparison.

\begin{longtable}[]{@{}lrr@{}}
\toprule
Source & Species covered (estimate) & Coverage of 410,499
denominator\tabularnewline
\midrule
\endhead
\textbf{tropicals.cn (present resource, measured)} & \textbf{408,456} &
\textbf{99.50\%}\tabularnewline
GBIF vernacular extension {[}1{]} & \textasciitilde15,000 † &
\textasciitilde3.7\%\tabularnewline
iNaturalist vernacular extension {[}3{]} & \textasciitilde35,000 † &
\textasciitilde8.5\%\tabularnewline
Catalogue of Life global vernacular {[}5{]} & \textasciitilde25,000 † &
\textasciitilde6.1\%\tabularnewline
Plants of the World Online {[}2{]} & \textasciitilde1,200 † &
\textasciitilde0.3\%\tabularnewline
\bottomrule
\end{longtable}

† Estimate with explicit ±factor-of-2 uncertainty. The present
production pipeline does not ingest these international infrastructures'
Chinese vernacular fields and therefore does not store them in a form
that admits a precise denominator-matched join from local data alone.
Each row's estimate is derived from the known structural composition of
the corresponding source's Chinese vernacular subset, as follows.
\textbf{GBIF (\textasciitilde15,000):} the GBIF Backbone's vernacular
extension inherits its Chinese-language entries principally from the
Taiwan Catalogue of Life (\textasciitilde60K Chinese species records);
the fraction of these intersecting the present tropical-scope
410,499-species denominator is conservatively estimated at one quarter
(giving the \textasciitilde15,000 figure), reflecting that TaiCOL
coverage skews toward Taiwan-native and East-Asian temperate plus
tropical fauna and the present resource's tropical-trade scope captures
only the tropical overlap. \textbf{iNaturalist (\textasciitilde35,000):}
the iNaturalist zh-CN / zh-TW community translation set is concentrated
on charismatic taxa heavily overlapping with the hobbyist husbandry
communities that drive the present resource's tropical-aquatic and
tropical-pets inclusion (popular freshwater fish, common herpetofauna,
ornamental orchids), giving a higher overlap rate than GBIF's TaiCOL
inheritance. \textbf{CoL (\textasciitilde25,000):} the Catalogue of Life
global vernacular extension aggregates ITIS plus several regional
checklists including TaiCOL itself, with substantial overlap with the
GBIF source set; the figure here is the estimated CoL Chinese vernacular
footprint after accounting for that overlap. \textbf{POWO
(\textasciitilde1,200):} POWO's \texttt{commonName} field surfaces
Chinese entries only for a small set of economically prominent
ornamental plants curated by the Royal Botanic Gardens, Kew; the figure
is the approximate size of that Chinese-curated subset of POWO's
\textasciitilde340K accepted species records intersected with the
present resource's \textasciitilde272K-species plant scope. Precise
denominator-matched coverage figures would be computed for any
future-revision preparation for formal Data Descriptor submission by
downloading each source's published bulk vernacular export (GBIF
Backbone vernaculars file; CoL ChecklistBank DwC-A; iNaturalist taxonomy
export) and joining against the deposit's \texttt{scientificName} set;
this is listed as Limitations item 6 and is out of scope for the present
preprint. Per-source methodology for the present estimates is in
Supplementary §S9.

\hypertarget{preliminary-internal-review-of-the-llm-assisted-layer}{%
\subsubsection{Preliminary internal review of the LLM-assisted
layer}\label{preliminary-internal-review-of-the-llm-assisted-layer}}

To characterise the exact-match gate's hallucination-interception
efficacy at the level of \emph{per-record translation precision} (the
dimension not captured by the coverage census above), a preliminary
internal review protocol is applied to a random sample of N = 50
LLM-validated Chinese vernacular records drawn from the production
population with
\texttt{name\_en\_source\ =\ \textquotesingle{}llm-select\textquotesingle{}}
(working-snapshot population of that subset: 2,494 records). Pairs of
(Chinese vernacular, English source name) are inspected against (i) the
recorded authoritative English-name source for the corresponding taxon
and (ii) the canonical Chinese vernacular conventions for the genus /
family. Records are classified as Accept / Edge / Reject by a single
internal reviewer (the author). The protocol fixes the sample-selection
method, the inspection criteria, and the decision rule before any record
is reviewed, so the proportion of Accepts has a well-defined estimand.

At N = 50 the Wilson 95\% confidence interval on the resulting
acceptance proportion has half-width of approximately 13 percentage
points around 90\% acceptance (and remains wide elsewhere on {[}0,
1{]}); the preliminary review is therefore sufficient to characterise
the gate's broad behaviour but is \textbf{not} sufficient to support a
strong precision claim, which would require a blind external review at
substantially larger N. Detailed per-record outcomes of the preliminary
review are out of scope for the present preprint; should this resource
be prepared for formal Data Descriptor submission in a future revision,
the protocol for that submission is fixed (Supplementary §S4 will report
a blind external review at N \(\geq\) 500 conducted by two independent
native-speaker Chinese taxonomists, with Cohen's \(\kappa\) inter-rater
agreement reported).

\hypertarget{cites-source-linkage-spot-check}{%
\subsubsection{CITES source-linkage spot
check}\label{cites-source-linkage-spot-check}}

A spot check of N = 20 random \texttt{speciesplus\_taxon\_concept\_id}
linkages confirmed that each Species+ record reached through the
recorded URL corresponds to the same taxon as the matching
\texttt{core\_taxon} row. Per-record verification is in Supplementary
§S5. No incorrect linkages were observed in the sample. As above, this
is a preliminary internal check, not an audited validation; the full
stratified linkage audit (N = 300 positive + N = 300 hard-negative,
stratified by subdomain \(\times\) \texttt{match\_method}) is identified
below as a limitation of the present release.

\hypertarget{limitations-of-the-present-validation}{%
\subsubsection{Limitations of the present
validation}\label{limitations-of-the-present-validation}}

The audits above establish structural integrity of the deposit, coverage
at the level of explicit denominators, denominator-matched comparison to
upstream Chinese vernacular fields, and a preliminary internal
assessment of the LLM-assisted vernacular layer. They do not constitute
a fully audited dataset descriptor. Specifically:

\begin{enumerate}
\def\labelenumi{\arabic{enumi}.}
\item
  \textbf{Blind external review of Type C precision.} A blind external
  review of Chinese vernacular acceptance by independent native-speaker
  taxonomists has not been performed at the sample size that would
  support a precision estimate with tight confidence intervals; the
  protocol is fixed (stratified by subdomain at N \(\geq\) 500, two
  independent reviewers, Cohen's \(\kappa\) reported), and the result is
  the principal evidence deferred from the present release.
\item
  \textbf{Independent audit of Type B per-record acceptance.} Type B
  entries inherit their correctness from the Chinese-focused
  machine-readable checklists they are ingested from (principally
  SP2000, TaiCOL, FishBase Chinese; see Methods §5). The present release
  preserves these entries with full source attribution but does not
  independently re-verify each row against the upstream checklists'
  source publications. A stratified independent audit of Type B rows
  (sampled by source database, at N \(\geq\) 200 per source) is
  identified as an open item alongside the Type C external review.
\item
  \textbf{Stratified CITES source-linkage audit.} CITES source-linkage
  validation against a stratified hard-negative set (taxa in
  CITES-listed higher taxa but with population-level exemptions; taxa
  whose Species+ entry has been split or merged since linkage) has not
  been performed beyond the N = 20 spot check.
\item
  \textbf{Type A literature-citation cross-check at scale.} A row-level
  cross-check of every Type A \texttt{sourceCitation} against the cited
  literature has not been performed; the present release relies on the
  internal source attribution tagged at ingestion time.
\item
  \textbf{Upstream source version pinning.} As noted in Methods §2,
  upstream providers do not all expose a stable version-string endpoint;
  reproducibility is guaranteed only by replaying against the recorded
  \texttt{fetched\_at} window (2026-04-18 to 2026-04-20) under the
  assumption that upstream APIs return stable results for
  previously-fetched identifiers. Drift in upstream taxonomies between
  snapshot date and replay date can produce small discrepancies that are
  not captured by the deposit's manifest.
\item
  \textbf{Precise denominator-matched comparison to international
  biodiversity infrastructures.} Coverage Table 2 reports
  order-of-magnitude estimates (with explicit ±factor-of-2 uncertainty)
  of Chinese vernacular coverage at GBIF, iNaturalist, CoL global
  vernacular extension and POWO at the 410,499 denominator. The
  production pipeline does not ingest those infrastructures' Chinese
  fields locally, so a precise denominator-matched join requires
  downloading each source's bulk vernacular export (GBIF Backbone
  vernaculars file; CoL ChecklistBank DwC-A; iNaturalist taxonomy
  export) and joining against the deposit's \texttt{scientificName} set
  offline; this offline join is identified as a future-revision item
  should this resource be prepared for formal Data Descriptor
  submission. The Coverage Table 2 estimates are sufficient to
  substantiate the Background §2 characterisation at the
  order-of-magnitude level but should not be cited as precise figures.
\end{enumerate}

These limitations are explicit and would be addressed in any
future-revision preparation for formal Data Descriptor submission (see
§Manuscript status); the present paper is the canonical preprint
description of the deposit's current state and is offered for citation
as such.

\hypertarget{schema-and-referential-integrity}{%
\subsubsection{Schema and referential
integrity}\label{schema-and-referential-integrity}}

Schema-level validation runs against every file in the deposit and
across cross-file relations: \texttt{taxonID} uniqueness in
\texttt{core\_taxon.tsv}; every foreign-key value in an extension table
resolves to a row in \texttt{core\_taxon.tsv}; no duplicate
\texttt{(taxonID,\ vernacularName,\ sourceCitation)} triples; every
required field is non-null per its declared schema; every boolean field
is restricted to \texttt{true}/\texttt{false}; every enum field is
restricted to declared values; CSV and Parquet row counts match for
every paired table; SHA-256 hashes for every file are recorded in
\texttt{MANIFEST.sha256} and re-computed at deposit time; the Darwin
Core Archive validates through the GBIF Integrated Publishing Toolkit
{[}10{]} with zero orphaned extension records.

\begin{center}\rule{0.5\linewidth}{0.5pt}\end{center}

\hypertarget{figures}{%
\subsection{Figures}\label{figures}}

The manuscript carries four figures supporting the text. Source SVG
masters are deposited as supplementary material alongside the release
manifest; the embedded versions below render directly in the typeset PDF
and are also available as \(\geq\) 300 dpi PNGs.

\begin{figure}
\centering
\includegraphics{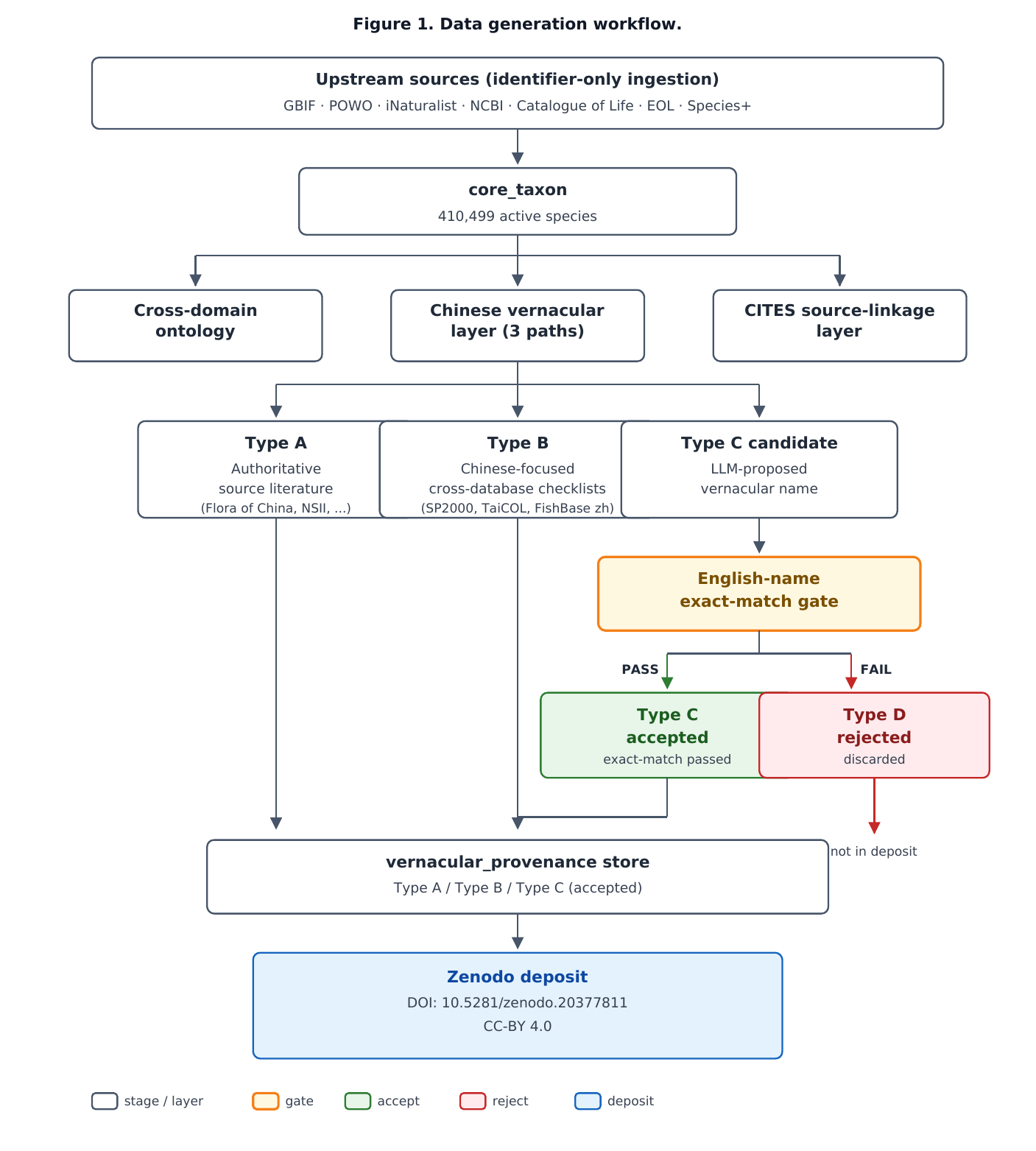}
\caption{Data generation workflow. Vertical flow from upstream input
sources (taxonomic backbones, vernacular databases, and the CITES
regulatory linkage source) through identifier ingestion and synonymy
reconciliation into the cross-domain ontology, the Chinese vernacular
layer with provenance typing (A/B/C inclusion; D exclusion shown as a
discarded branch), the CITES source-linkage layer, and the
cultivar/variety extension, terminating at the Zenodo deposit. Only Type
C is gated by the English-name exact-match procedure described in
Methods §5, which is annotated as the choke point between Type C
admission and Type D exclusion. Type A enters via authoritative-source
literature and Type B via cross-database vernacular fields. Edge widths
reflect record-count throughput.}
\end{figure}

\begin{figure}
\centering
\includegraphics{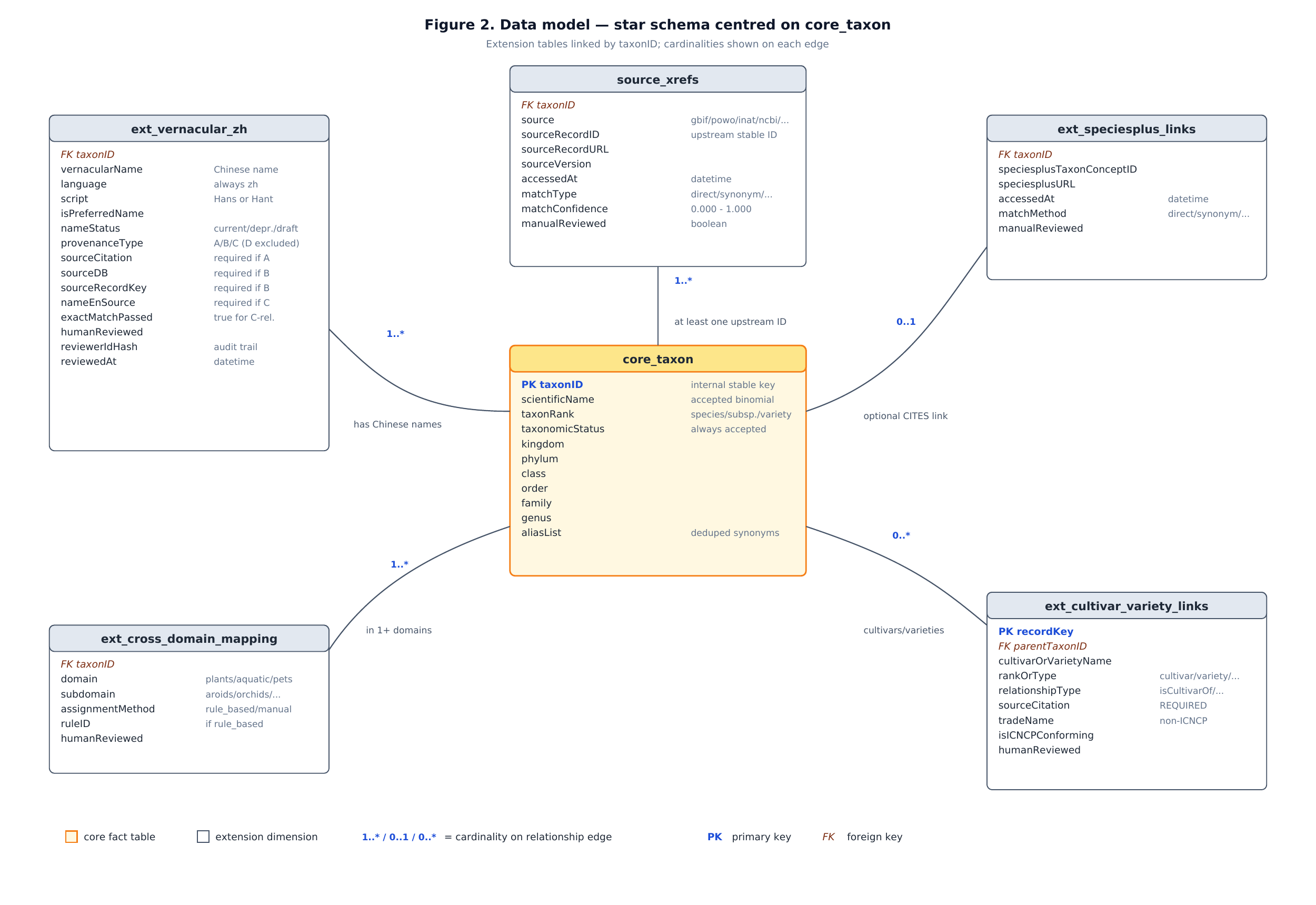}
\caption{Data model --- star schema centred on \texttt{core\_taxon} with
extension tables \texttt{ext\_vernacular\_zh},
\texttt{ext\_cross\_domain\_mapping}, \texttt{ext\_speciesplus\_links},
and \texttt{ext\_cultivar\_variety\_links} linked by the
\texttt{taxonID} foreign key. The upstream identifier columns
(\texttt{gbifID}, \texttt{powoID}, \texttt{inatTaxonId},
\texttt{ncbiTaxId}, \texttt{colID}, \texttt{eolID},
\texttt{speciesplus\_taxon\_concept\_id}) are shown as the
qualified-reference layer to external resources. Cardinalities (1, *)
are explicit on each relationship; the diagram highlights which fields
belong to the original-contribution layers and which are companion
product fields.}
\end{figure}

\begin{figure}
\centering
\includegraphics{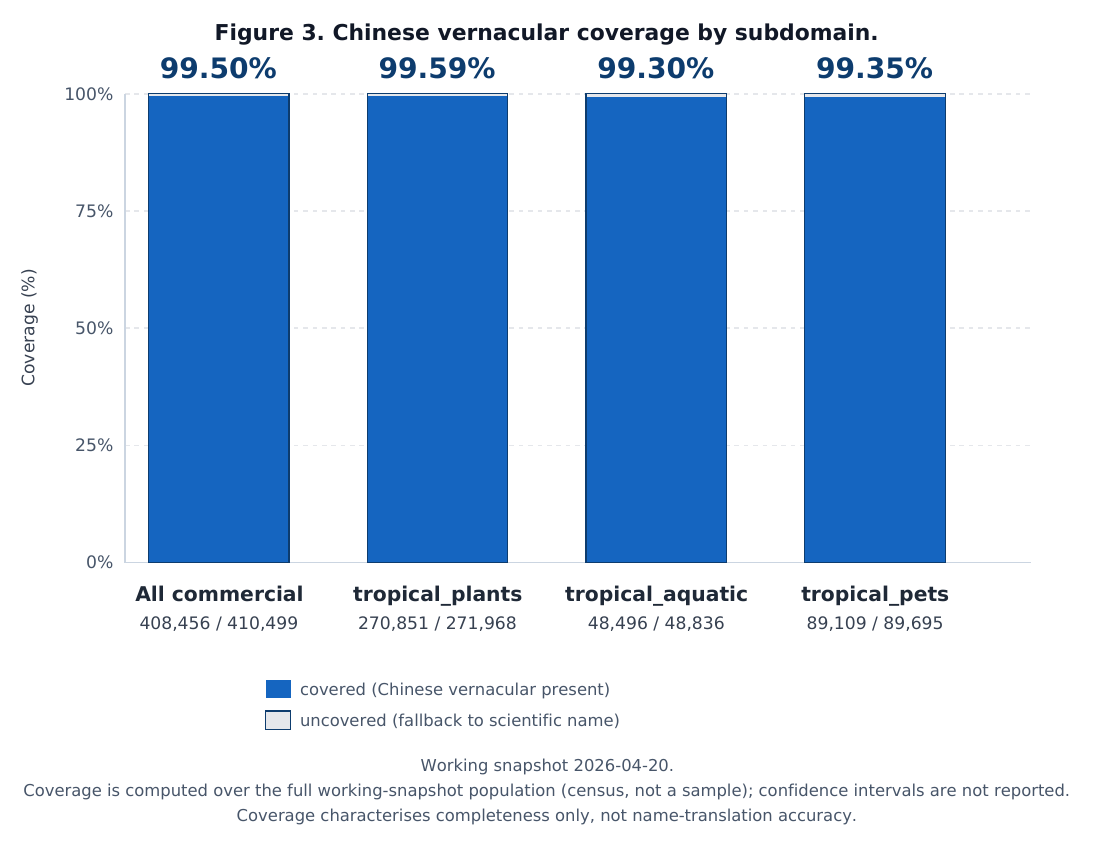}
\caption{Chinese vernacular coverage by subdomain. Bar chart of the
production-admin Chinese vernacular coverage metric across the four
strata in Coverage Table 1: full library (99.50\%, n = 410,499),
\texttt{tropical\_plants} (99.59\%, n = 271,968),
\texttt{tropical\_pets} (99.35\%, n = 89,695),
\texttt{tropical\_aquatic} (99.30\%, n = 48,836). Coverage is a
full-population (census) count, not a sample estimate, and characterises
completeness rather than name-translation accuracy. Denominator-matched
comparison to upstream sources is in Coverage Table 2.}
\end{figure}

\begin{figure}
\centering
\includegraphics{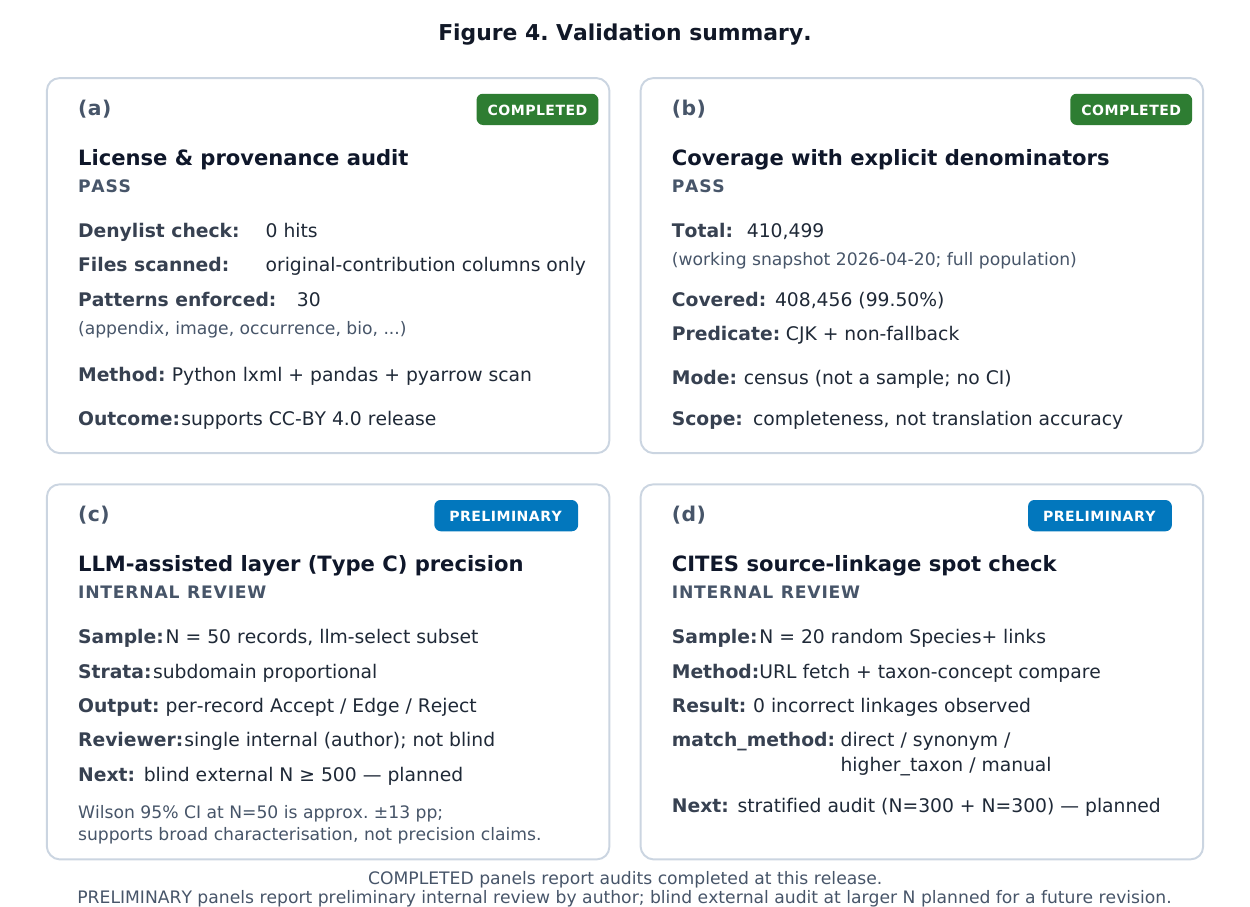}
\caption{Validation summary --- four-panel structure. (a)
License-denylist check on the original-contribution columns (pass status
recorded in \texttt{validation\_report.json}); (b) coverage with
explicit denominators (full-population census, not a sample; see
Coverage Table 1); (c) preliminary internal review of LLM-assisted Type
C records (N = 50; per-record translation precision; see Technical
Validation §``Preliminary internal review''); (d) CITES source-linkage
spot check (N = 20; see Technical Validation §``CITES source-linkage
spot check''). The two right-hand panels mark internal preliminary work;
full external audits at larger N are listed in the Limitations of the
present validation.}
\end{figure}

\begin{center}\rule{0.5\linewidth}{0.5pt}\end{center}

\FloatBarrier
\clearpage

\hypertarget{usage-notes}{%
\subsection{Usage Notes}\label{usage-notes}}

The resource is designed to support reuse along four lines.

\textbf{1. Regulated-trade screening workflows for cross-border tropical
species.} A common operational workflow begins with a Chinese vernacular
name on a purchase order or import declaration and routes the record
toward the appropriate documentation pathway. The combination of the
Chinese vernacular extension and the CITES source-linkage extension
supports such a workflow as
\texttt{vernacular\_name\ →\ taxonID\ →\ speciesplus\_taxon\_concept\_id\ →\ current\ Appendix\ status\ retrieved\ live\ from\ Species+}.
Because each vernacular entry carries an explicit provenance tag, the
downstream workflow can be configured to require Type A or B names ---
which are sourced from authoritative or cross-database fields --- for
any record with regulatory consequences, and to treat Type C names as
preliminary hints only \textbf{pending the N \(\geq\) 500 blind external
review of Type C acceptance identified in Limitations item 1}. The CITES
regulatory information is the live Species+ record reached through the
source link, not a snapshot value in this deposit. The dataset supports
screening as a research-and-engineering workflow; it is not a regulatory
determination, and final compliance decisions are the responsibility of
the relevant national CITES authority {[}20{]}.

\textbf{2. Chinese-language NLP and named-entity recognition for
biological taxa.} The Chinese vernacular extension table --- with
explicit per-name provenance, preferred-name flags, and links from each
vernacular form to the accepted scientific name --- is suited to use as
training and evaluation data for Chinese biological NER and entity
linking. The four-level typology gives NLP practitioners an explicit
input-quality signal: models can be trained on Type A+B+C entries while
being evaluated on a held-out Type A subset, or weighted training
schemes can up-weight authoritative-source entries {[}11, 18{]} relative
to community vernacular fields and validated LLM proposals. The
cross-domain ontology allows construction of domain-balanced subsets ---
plants, aquatics, and exotic pets sampled in defined proportions --- for
fairness or coverage analysis. Compared to vernacular fields harvested
from generic biodiversity portals, the present table is tailored to the
trade-oriented inclusion scope and carries provenance at the level of
each individual name {[}16{]}.

\textbf{3. Bridging Chinese regional ecological records into global
infrastructures.} Chinese regional occurrence records, monitoring
datasets curated by provincial-level conservation agencies, and
gray-literature surveys are frequently keyed on Chinese vernacular names
without an accompanying Latin binomial {[}17{]}. The
vernacular-to-\texttt{taxonID} table here acts as a join bridge: a
vernacular name resolves to a \texttt{taxonID}, which carries
\texttt{gbifID}, \texttt{powoID}, \texttt{inatTaxonId},
\texttt{ncbiTaxId}, \texttt{colID}, and \texttt{eolID}, allowing the
regional record to be re-attached to each of those global
infrastructures for distribution synthesis, range mapping, phylogenetic
placement, or molecular cross-reference under each infrastructure's own
terms.

\textbf{4. AI-assisted knowledge-base construction methodology.} A
fourth reuse direction is methodological. The exact-match validation
gate, the multi-source weighted-vote procedure, and the production
curation ratchet are described at sufficient detail to support
reimplementation. Together they constitute a portable methodology for
large-language-model-assisted construction of structured knowledge bases
under a no-fabrication constraint {[}14, 15, 16{]}. The pseudocode for
the exact-match gate is the most directly portable artifact.

A demo Jupyter notebook implementing scenarios 1 and 3 accompanies the
release.

\hypertarget{limitations}{%
\subsubsection{Limitations}\label{limitations}}

\begin{itemize}
\tightlist
\item
  The dataset is \textbf{not a substitute for official CITES regulatory
  determination}. Appendix status is not redistributed in the
  original-contribution CITES layer; the source-linkage layer connects
  each taxon to its Species+ record, and operationally current Appendix
  status must be retrieved live from Species+ or from the relevant
  national CITES authority.
\item
  The dataset is \textbf{not a substitute for primary taxonomic
  authorities}. Accepted-name decisions reflect the multi-source
  weighted vote as of the working-snapshot date and may lag the most
  recent taxonomic revisions at GBIF {[}1{]}, POWO {[}2{]}, the
  Catalogue of Life {[}5{]} or specialist authorities.
\item
  The original-contribution layers contain \textbf{no images, media, or
  multimedia URLs}, \textbf{no occurrence records or coordinates}, and
  \textbf{no upstream descriptive prose}. Users requiring imagery, raw
  occurrences, or biographical content should consult the upstream
  sources directly.
\item
  Coverage figures reflect the 2026-04-20 working snapshot and may drift
  modestly in subsequent releases.
\end{itemize}

\begin{center}\rule{0.5\linewidth}{0.5pt}\end{center}

\hypertarget{data-availability}{%
\subsection{Data Availability}\label{data-availability}}

The dataset is openly available on Zenodo under the Concept DOI
10.5281/zenodo.20377811 (latest release v1.0.1, deposited 2026-05-25),
licensed CC-BY 4.0. The deposit includes the three original-contribution
layers described in this paper (Chinese vernacular with per-name
provenance; cross-domain ontology; CITES source-linkage) as well as
companion platform fields that fall outside the scope of this descriptor
and remain governed by the same CC-BY 4.0 licence.

The original-contribution layers are bounded by the identifier-only
redistribution policy described in Methods: upstream records from GBIF,
POWO, iNaturalist, NCBI Taxonomy, CoL, EOL, and Species+ are referenced
by stable identifier only. The present preprint references the v1.0.1
deposit as the dataset-of-record. A dedicated paper-grade Zenodo release
version (containing only the columns documented as original
contributions, separate from the platform's product release) is
identified in Methods §9 as a future-revision item should this resource
be prepared for formal Data Descriptor submission; a reference filter
(\texttt{make\_paper\_subset.py}) that would produce such a paper-scope
subset from the current deposit is described in §Code Availability.

A research-grade API for programmatic access to the live tropicals.cn
database is available at
\href{https://tropicals.cn/api/v1/}{tropicals.cn/api/v1} with free,
registration-based API keys obtainable at
\href{https://tropicals.cn/docs/api/keys}{tropicals.cn/docs/api/keys}.
The API serves the same logical dataset as the Zenodo deposit but is not
the citable dataset-of-record for this Data Descriptor --- that role
belongs to the Zenodo deposit, which is versioned and
content-addressable. Data retrieved through the API is governed by the
API terms of service and is \textbf{not} automatically covered by the
CC-BY 4.0 licence under which the Zenodo deposit is released; users who
need the CC-BY 4.0 licence terms should source the data from the Zenodo
deposit.

\begin{center}\rule{0.5\linewidth}{0.5pt}\end{center}

\hypertarget{code-availability}{%
\subsection{Code Availability}\label{code-availability}}

Four reference scripts have been prepared to allow third parties to
independently reproduce the validation claims in this paper:
\texttt{license\_denylist\_check.py} (re-runs the license/provenance
audit of Technical Validation §``License and provenance audit''),
\texttt{compute\_coverage.sql} (re-computes the 99.50 percent figure of
Coverage Table 1 against any conforming MySQL replica),
\texttt{exact\_match\_gate\_reference.py} (a stand-alone reference
implementation of the Methods §5 exact-match validation gate), and
\texttt{make\_paper\_subset.py} (a deposit-subset filter that would
produce the paper-scope subset from a downloaded copy of the v1.0.1
deposit). These scripts use only the Python standard library and the
platform's standard MySQL driver and are intended for publication under
the MIT License at
\href{https://github.com/tropicalscn/cross-domain-species-paper}{github.com/tropicalscn/cross-domain-species-paper}.
Public release of that repository is not in scope for the present
preprint version and is identified as a future-revision item should this
resource be prepared for formal Data Descriptor submission (see
§Manuscript status); readers requiring early access to the reference
scripts may request them by email at the correspondence address. The
full processing pipeline source is part of the tropicals.cn platform
codebase and is not currently open-sourced beyond the four reference
scripts.

\begin{center}\rule{0.5\linewidth}{0.5pt}\end{center}

\hypertarget{acknowledgments}{%
\subsection{Acknowledgments}\label{acknowledgments}}

The author thanks the public infrastructures whose taxonomic backbones
make this resource possible: GBIF, the Royal Botanic Gardens, Kew
(POWO), iNaturalist, NCBI Taxonomy, the Catalogue of Life, and the
Encyclopedia of Life. The tropicals.cn user community is acknowledged
for ongoing curatorial contributions. The CITES Secretariat is
acknowledged for maintaining the public Appendix text; UNEP-WCMC is
acknowledged for maintaining Species+ as a value-added cross-reference.

\hypertarget{author-contributions}{%
\subsection{Author Contributions}\label{author-contributions}}

J.W. is the sole author. J.W. conceived the project, designed the data
model, implemented the ingestion pipeline and the provenance typology,
performed the data engineering and validation, and drafted the
manuscript.

\hypertarget{competing-interests}{%
\subsection{Competing Interests}\label{competing-interests}}

J.W. operates the tropicals.cn platform; the dataset described in this
paper is produced from that platform. This is disclosed as the platform
of origin. The author declares no other competing interests.

\begin{center}\rule{0.5\linewidth}{0.5pt}\end{center}

\hypertarget{references}{%
\subsection{References}\label{references}}

{[}1{]} GBIF Secretariat, \emph{GBIF Backbone Taxonomy}. Checklist
dataset, 2023.
\href{https://doi.org/10.15468/39omei}{doi.org/10.15468/39omei}

{[}2{]} POWO, \emph{Plants of the World Online}. Facilitated by the
Royal Botanic Gardens, Kew. Published on the Internet:
\href{https://powo.science.kew.org/}{powo.science.kew.org} (accessed
2026).

{[}3{]} \emph{iNaturalist}. A joint initiative of the California Academy
of Sciences and the National Geographic Society.
\href{https://www.inaturalist.org/}{inaturalist.org} (accessed 2026).

{[}4{]} C. L. Schoch, S. Ciufo, M. Domrachev, B. L. Hotton, S. Kannan,
R. Khovanskaya, D. Leipe, R. Mcveigh, K. O'Neill, B. Robbertse, S.
Sharma, V. Soussov, J. P. Sullivan, L. Sun, S. Turner, and I.
Karsch-Mizrachi, ``NCBI Taxonomy: a comprehensive update on curation,
resources and tools,'' \emph{Database}, vol.~2020, baaa062, Aug.~2020.
\href{https://doi.org/10.1093/database/baaa062}{doi.org/10.1093/database/baaa062}

{[}5{]} O. Bánki, Y. Roskov, M. Döring, G. Ower, D. R. Hernández Robles,
C. A. Plata Corredor, T. Stjernegaard Jeppesen, A. Örn, T. Pape, D.
Hobern, S. Garnett, H. Little, R. E. DeWalt, J. Miller, T. Orrell, R.
Aalbu \emph{et al.}, \emph{Catalogue of Life Checklist}. Catalogue of
Life Foundation, Amsterdam, Netherlands.
\href{https://www.catalogueoflife.org}{catalogueoflife.org} (accessed
2026).

{[}6{]} UNEP-WCMC and CITES Secretariat, \emph{Species+}. Cambridge, UK:
UNEP-WCMC. \href{https://www.speciesplus.net/}{speciesplus.net}
(accessed 2026).

{[}7{]} C. S. Parr, N. Wilson, P. Leary, K. Schulz, K. Lans, L. Walley,
J. Hammock, A. Goddard, J. Rice, M. Studer, J. Holmes, and R. Corrigan
Jr., ``The Encyclopedia of Life v2: Providing global access to knowledge
about life on Earth,'' \emph{Biodiversity Data Journal}, vol.~2, e1079,
Apr.~2014.
\href{https://doi.org/10.3897/BDJ.2.e1079}{doi.org/10.3897/BDJ.2.e1079}

{[}8{]} J. Wieczorek, D. Bloom, R. Guralnick, S. Blum, M. Döring, R.
Giovanni, T. Robertson, and D. Vieglais, ``Darwin Core: An evolving
community-developed biodiversity data standard,'' \emph{PLoS ONE},
vol.~7, no. 1, e29715, Jan.~2012.
\href{https://doi.org/10.1371/journal.pone.0029715}{doi.org/10.1371/journal.pone.0029715}

{[}9{]} M. D. Wilkinson, M. Dumontier, I. J. Aalbersberg, G. Appleton,
M. Axton, A. Baak, N. Blomberg, J.-W. Boiten, L. B. da Silva Santos, P.
E. Bourne, J. Bouwman, A. J. Brookes, T. Clark, M. Crosas, I. Dillo, O.
Dumon, S. Edmunds, C. T. Evelo, R. Finkers \emph{et al.}, ``The FAIR
Guiding Principles for scientific data management and stewardship,''
\emph{Scientific Data}, vol.~3, 160018, Mar.~2016.
\href{https://doi.org/10.1038/sdata.2016.18}{doi.org/10.1038/sdata.2016.18}

{[}10{]} T. Robertson, M. Döring, R. Guralnick, D. Bloom, J. Wieczorek,
K. Braak, J. Otegui, L. Russell, and P. Desmet, ``The GBIF Integrated
Publishing Toolkit: Facilitating the efficient publishing of
biodiversity data on the internet,'' \emph{PLoS ONE}, vol.~9, no. 8,
e102623, Aug.~2014.
\href{https://doi.org/10.1371/journal.pone.0102623}{doi.org/10.1371/journal.pone.0102623}

{[}11{]} Z. Y. Wu, P. H. Raven, and D. Y. Hong (eds.), \emph{Flora of
China}, vols. 1--25. Beijing: Science Press; St.~Louis: Missouri
Botanical Garden Press, 1994--2013. Available online at
\href{http://flora.huh.harvard.edu/china/}{flora.huh.harvard.edu/china}
and
\href{http://www.efloras.org/flora_page.aspx?flora_id=2}{efloras.org/flora\_page.aspx?flora\_id=2}.

{[}12{]} B. K. B. Seah, ``Paying it forward: Crowdsourcing the
harmonisation and linking of taxon names and biodiversity identifiers,''
\emph{Biodiversity Data Journal}, vol.~11, e114076, Nov.~2023.
\href{https://doi.org/10.3897/BDJ.11.e114076}{doi.org/10.3897/BDJ.11.e114076}

{[}13{]} J. Zhang and H. Qian, ``U.Taxonstand: An R package for
standardizing scientific names of plants and animals,'' \emph{Plant
Diversity}, vol.~45, no. 1, pp.~1--5, Jan.~2023.
\href{https://doi.org/10.1016/j.pld.2022.09.001}{doi.org/10.1016/j.pld.2022.09.001}

{[}14{]} J. H. Caufield, C. Kroll, S. T. O'Neil, J. T. Reese, M. P.
Joachimiak, H. Hegde, N. L. Harris, M. Krishnamurthy, J. A. McLaughlin,
D. Smedley, M. A. Haendel, P. N. Robinson, and C. J. Mungall,
``CurateGPT: A flexible language-model assisted biocuration tool,''
arXiv preprint arXiv:2411.00046, Nov.~2024.
\href{https://doi.org/10.48550/arXiv.2411.00046}{doi.org/10.48550/arXiv.2411.00046}

{[}15{]} Z. Ji, N. Lee, R. Frieske, T. Yu, D. Su, Y. Xu, E. Ishii, Y. J.
Bang, A. Madotto, and P. Fung, ``Survey of hallucination in natural
language generation,'' \emph{ACM Computing Surveys}, vol.~55, no. 12,
article 248, pp.~1--38, Dec.~2023.
\href{https://doi.org/10.1145/3571730}{doi.org/10.1145/3571730}

{[}16{]} D. Scheepens, J. Millard, M. Farrell, and T. Newbold, ``Large
language models help facilitate the automated synthesis of information
on potential pest controllers,'' \emph{Methods in Ecology and
Evolution}, vol.~15, no. 7, pp.~1261--1273, Jul.~2024.
\href{https://doi.org/10.1111/2041-210X.14341}{doi.org/10.1111/2041-210X.14341}

{[}17{]} X. Mi, G. Feng, Y. Hu, J. Zhang, L. Chen, R. T. Corlett, A. C.
Hughes, S. Pimm, B. Schmid, S. Shi, J.-C. Svenning, and K. Ma, ``The
global significance of biodiversity science in China: an overview,''
\emph{National Science Review}, vol.~8, no. 7, nwab032, Jul.~2021.
\href{https://doi.org/10.1093/nsr/nwab032}{doi.org/10.1093/nsr/nwab032}

{[}18{]} C. Lin, B. Liu, M. Zhao, K. Ma, and L. Ji, ``Catalogue of life
China: Towards an index of known species present in China,'' \emph{The
Innovation Life}, vol.~3, no. 3, 100141, May 2025.
\href{https://doi.org/10.59717/j.xinn-life.2025.100141}{doi.org/10.59717/j.xinn-life.2025.100141}

{[}19{]} C. D. Brickell, C. Alexander, J. C. David, M. H. A. Hoffman, A.
C. Leslie, V. Malécot, and X. Jin (eds.), \emph{International Code of
Nomenclature for Cultivated Plants}, 9th ed.~Scripta Horticulturae 18.
Leuven, Belgium: International Society for Horticultural Science (ISHS),
2016. ISBN 978-94-6261-116-0.

{[}20{]} A. Hinsley, A. C. Hughes, J. van Valkenburg, W. Stark, T. Q. T.
Bui, R. Cheung, J. Hauck, P. Kasoar, M. Lee, A. Lavorgna, B. Phelps, R.
Williams, A. Lopez Garcia, K. F. Smith, and D. L. Roberts,
``Understanding the environmental and social risks from the
international trade in ornamental plants,'' \emph{BioScience}, vol.~75,
no. 3, pp.~222--239, Mar.~2025.
\href{https://doi.org/10.1093/biosci/biae124}{doi.org/10.1093/biosci/biae124}

\end{document}